\documentclass[10pt,twocolumn,letterpaper]{article}

\usepackage{wacv}
\usepackage{times}
\usepackage{epsfig}
\usepackage{graphicx}
\usepackage{amsmath}
\usepackage{amssymb}
\usepackage{booktabs}
\usepackage{multirow}
\usepackage{subfig}

\wacvfinalcopy %

\def\wacvPaperID{42} %

\ifwacvfinal\fi
\begin{document}

\newcommand\blfootnote[1]{
  \begingroup
  \renewcommand\thefootnote{}\footnote{#1}
  \addtocounter{footnote}{-1}
  \endgroup
}

\title{It's All About The Scale - Efficient Text Detection Using Adaptive Scaling}

\author{Elad Richardson\textsuperscript{*}, Yaniv Azar, Or Avioz, Niv Geron,  Tomer Ronen, Zach Avraham, Stav Shapiro   \\
Penta-AI\\
}

\maketitle
\ifwacvfinal\thispagestyle{empty}\fi

\begin{abstract}
``Text can appear anywhere''. This property requires us to carefully process all the pixels in an image in order to accurately localize all text instances. In particular, for the more difficult task of localizing small text regions, many methods use an enlarged image or even several rescaled ones as their input. This significantly increases the processing time of the entire image and needlessly enlarges background regions.
If we were to have a prior telling us the coarse location of text instances in the image and their approximate scale, we could have adaptively chosen which regions to process and how to rescale them, thus significantly reducing the processing time.
To estimate this prior we propose a segmentation-based network with an additional ``scale predictor'', an output channel that predicts the scale of each text segment. The network is applied on a scaled down image to efficiently approximate the desired prior, without processing all the pixels of the original image.
 The approximated prior is then used to create a compact image containing \textbf{only text regions}, resized to a \textbf{canonical scale}, which is fed again to the segmentation network for fine-grained detection. We show that our approach offers a powerful alternative to fixed scaling schemes, achieving an equivalent accuracy to larger input scales while processing far fewer pixels. Qualitative and quantitative results are presented on the ICDAR15 and ICDAR17 MLT benchmarks to validate our approach.
\end{abstract}

\section{Introduction}
\begin{figure}
    \centering
    \includegraphics[clip, trim=0cm 12cm 0cm 0cm,width=0.48\textwidth]{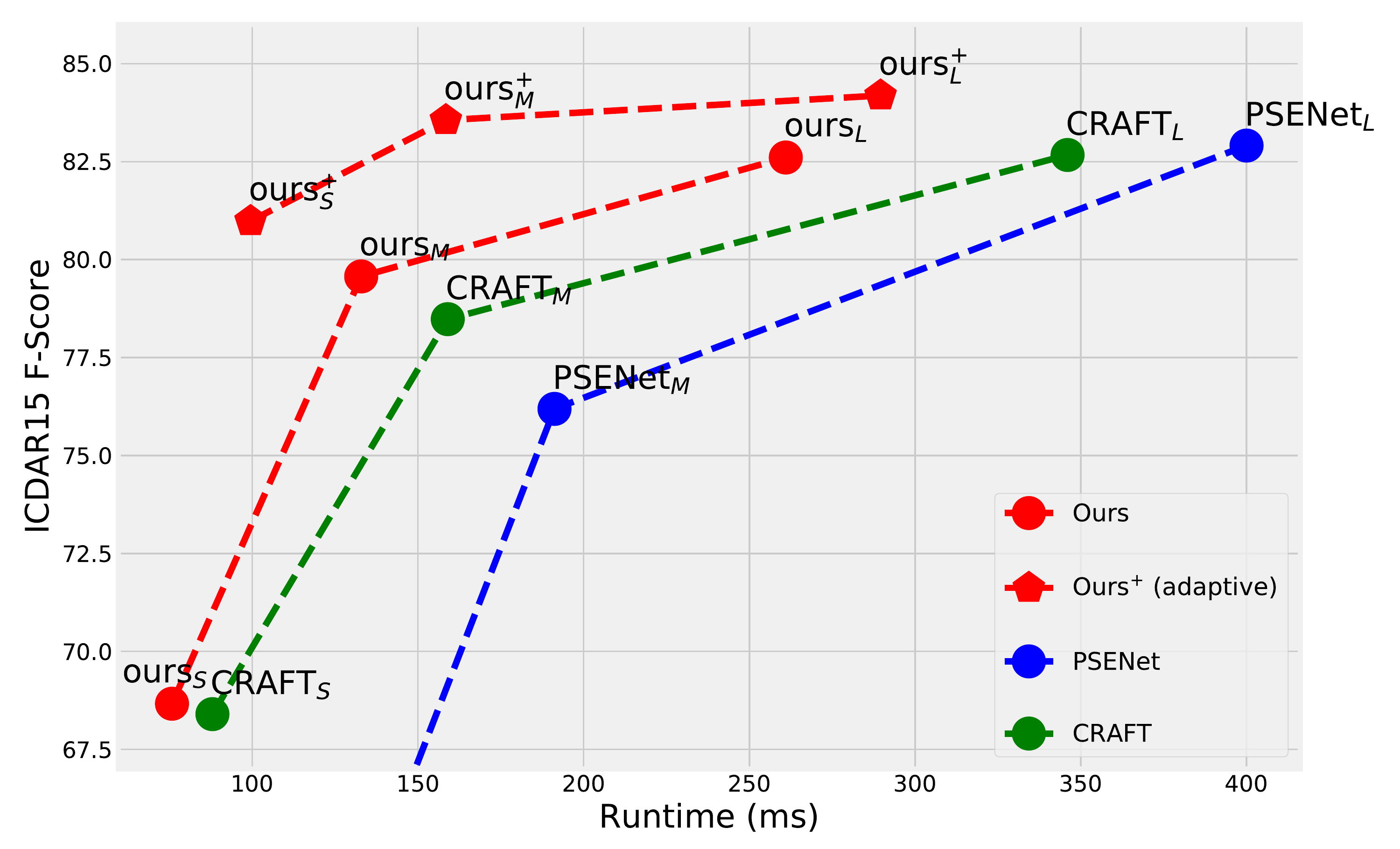}
    \caption{The ``performance vs accuracy'' trade-off. $S$, $M$ and $L$ denote an input with a long side of $720$, $1024$ and $1440$ accordingly. Ours is the proposed single-scale method, while $\mbox{Ours}^{+}$ is boosted using the proposed scaling scheme.}
\label{fig:results_graph}
\end{figure}

\begin{figure*}
  \centering
 \includegraphics[width=1.0\linewidth]
     {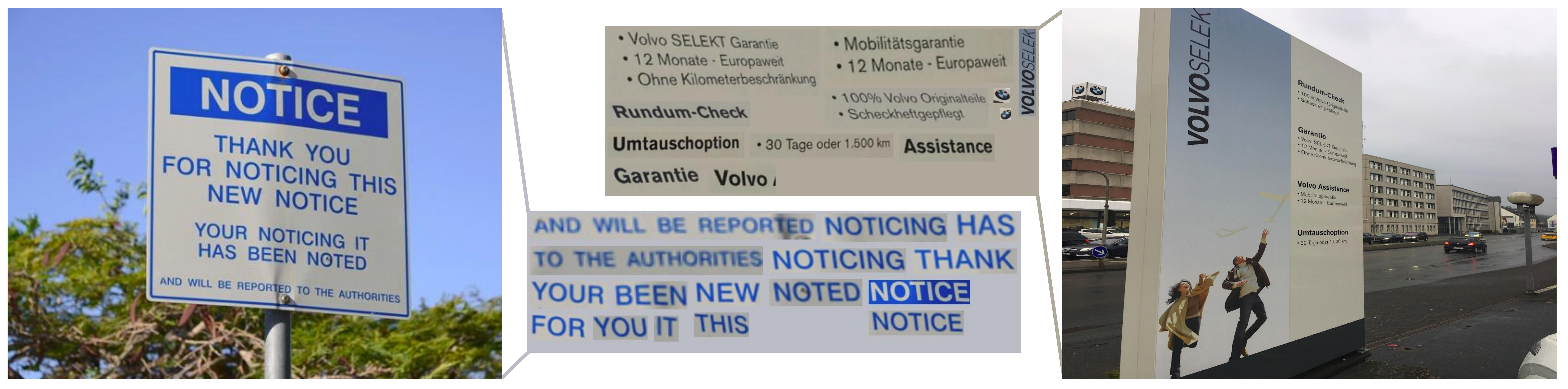}\captionof{figure}{Canonical representations, ``knapsacks'', created using the proposed approach on ICDAR17 images.  In the left image, note how scaling is applied adaptively to create a uniform scale. In the right image, see how all background regions are removed before rescaling.
     }\label{fig:compacts}
\end{figure*}

Reading text from natural images is a long-standing problem in the field of computer vision.
Usually, the problem involves two stages: (1) A text detection mechanism, whose purpose is to localize the individual words in the image, and (2) A text recognition mechanism, whose purpose is to take each detected text region and parse it into a single word. \blfootnote{* Corresponding author. Email: elad.richardson@gmail.com} \

When it comes to the detection stage, recent methods have made impressive leaps in terms of performance~\cite{he2017single,liao2017textboxes,liao2018textboxes++,deng2018pixellink,zhou2017east,wang2019shape,baek2019character,liu2018fots,liao2018rotation}. These methods can most often be classified into two distinct types. The first type is anchor-based approaches~\cite{he2017single,liao2017textboxes,liao2018textboxes++,liu2018fots,liao2018rotation}, which build upon popular object detection CNN architectures, such as SSD~\cite{liu2016ssd}, Yolo~\cite{redmon2016you}, or Faster R-CNN~\cite{ren2015faster}, and directly predict a bounding box or quadrilateral around the text. While efficient, they are less suited for detecting rotated or irregular text.
The second type is segmentation-based methods~\cite{deng2018pixellink,zhou2017east,wang2019shape,baek2019character}, which usually predict, for each pixel, a text/no-text semantic mask from which bounding boxes are extracted using an additional post-processing stage. While this representation is more flexible, it struggles with small text instances which are close to one another and cannot be easily separated.

Indeed, one of the major difficulties in text detection in general, and specifically with the segmentation approaches, lies in detecting small text instances. While much effort has been put into the problem by creating better post-processing schemes~\cite{deng2018pixellink,wang2019shape,long2018textsnake},
the problem of finding better scaling schemes is somewhat overlooked. Instead, most methods simply resort to fixed scaling schemes which are applied on top of their proposed baseline. That is, feeding the same image into the baseline network in an enlarged scale, or even multiple ones.
Although effective in terms of recall, these schemes are wasteful both in terms of runtime and memory. This can be attributed to two main factors:
\begin{enumerate}
  \item In many cases, \textbf{text occupies only small regions of the original image}. Thus, when enlarging the image to capture small text instances we also redundantly process many more background pixels.
  \item  Enlarging the entire image changes the scale of all text regions, even though \textbf{many text regions might already be in an appropriate scale for detection} and do not need to be enlarged. Large text might even be harder to detect when further enlarged. To mitigate that some methods choose to run multiple fixed scales, but this also increases the processing cost.
\end{enumerate}

In this paper, an approach is presented to tackle these problems. The core idea behind our approach is that localizing regions of text is much easier than localizing individual words.
Hence, we propose to utilize a coarse forward evaluation on a downsized image to locate text regions while simultaneously approximating the scale of each such region. This information is used to create a compact representation containing \textbf{only text regions}, where each region is resized to a \textbf{canonical scale}, as shown in Figure~\ref{fig:compacts}. The compact representation can then be processed using a single forward pass, resulting in a much more efficient evaluation process.

\begin{figure*}
    \centering
    \includegraphics[width=0.98\textwidth]{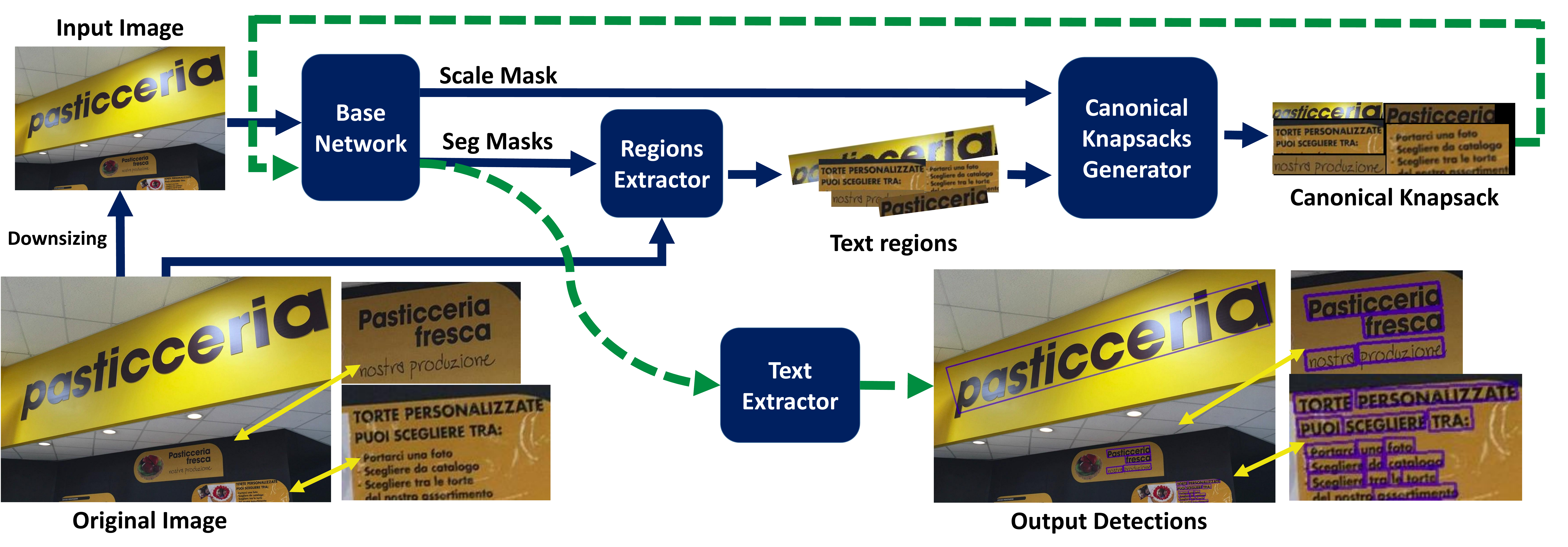}
    \caption{The proposed pipeline. First, a downsized image is fed into our base network to get initial segmentation and scale masks. These masks are then used to create a canonical knapsack, containing only text regions in a uniform scale. This knapsack is then fed again through the baseline network where the segmentation mask is then used to create the refined localization of text instances.}
\label{fig:pipeline}
\end{figure*}

In practice, this is achieved by taking a semantic segmentation method and adding an output channel that represents the height of each text instance. While this information is redundant when the detections are well separated, it is crucial when several text instances are merged. That is because the raw segmentation mask cannot tell us whether a text region contains a single large-scale line or several merged lines of smaller text.  The height channel, however, allows us to easily retrieve the scale of each such region and scale it as needed, assuring that the text will be well separated in the second pass. These scaled text regions are then packed together into a single image, or ``knapsack'', that is fed again into the same segmentation network. The full process is shown in Figure~\ref{fig:pipeline}.

To validate our approach we propose and implement a new semantic segmentation baseline, based on recent state-of-the-art approaches, which uses a simple and efficient post-processing scheme. Our scale channel is added to the proposed baseline to create the final network. The approach is validated under varying conditions and benchmarks, showing that our adaptive method is indeed a powerful alternative to fixed scaling schemes.

\noindent The main contributions of this paper are:

\begin{itemize}
    \item A novel scheme for adaptively scaling text images resulting in a far more efficient process compared to fixed scaling schemes.
    \item An improved semantic segmentation approach for text detection requiring only a simple post-processing step.
\end{itemize}

\section{Related Work}
As mentioned above, current methods can be roughly divided to anchor-based approaches~\cite{he2017single,liao2017textboxes,liao2018textboxes++,liu2018fots,ma2018arbitrary,liao2018rotation} and segmentation-based ones~\cite{deng2018pixellink,zhou2017east,wang2019shape,he2016accurate,yao2016scene,zhang2016multi}, where some recent methods try to fuse the two types together~\cite{li2018pixel,liu2019pyramid,huang2019mask,lyu2018mask}.
Our proposed pipeline is based on recent segmentation-based text detection methods, which are discussed next. Details on other approaches which are not covered in this work are presented in~\cite{long2018scene}.

Segmentation-based text detection approaches have gained significant attention in recent years, starting from the seminal works of Yao \etal{}~\cite{yao2016scene} and Zhang \etal{}~\cite{zhang2016multi}. These works solve the problem of text detection by reformulating it as a semantic segmentation scheme, which is then solved by a Fully Convolutional Network (FCN)~\cite{long2015fully}.
It was shown that these approaches are better suited for rotated and irregular text and interest in them has subsequently emerged.
These methods, however, share a common problem where adjacent word instances tend to connect.
This problem is inherent in nearly all segmentation based methods, and recent segmentation based approaches for text detection put a large emphasis on mitigating this problem~\cite{polzounov2017wordfence,deng2018pixellink,wang2019shape,baek2019character}.

The WordFence approach~\cite{polzounov2017wordfence} learns an additional border class to force a better separation of word instances.
PixelLink~\cite{deng2018pixellink} tries to predict, for each pixel in an 8-connected neighborhood, whether its neighbors belong to the same text label. The predicted connectivity maps, in addition to the original text/no-text segmentation map, are then used to generate the final detections.
The recent PSENet method~\cite{wang2019shape} learns a set of scaled kernels around each text instance, which, in test time, are progressively expanded to generate the complete word instance prediction. The CRAFT method~\cite{baek2019character} uses character affinity maps to connect character detections into a single word.
While both PSENet and CRAFT achieve state-of-the-art results on several competitive benchmarks they require extremely large input images. For example, on the ICDAR15 benchmark~\cite{karatzas2015icdar}, images are enlarged from $720\times1280$ to $1260\times2240$, which significantly increases runtime and can present difficulties on platforms with limited resources.
While some approaches have tried to apply a two-stage approach, in which rough text regions are first located~\cite{he2016accurate,yue2018boosting,he2019textcontournet}, to the best of our knowledge we are the first to directly learn a text scale channel in order to build an optimized two-stage detection pipeline.

In contrast to previously mentioned works, research into adaptive scaling schemes for object and text instance detection is far less prominent. In~\cite{chin2019adascale} it is shown that by learning a single optimal scale for each image, it is possible to improve both the accuracy and speed of object detection. Yuan \etal{}~\cite{yuan2018scale} learns scale-adaptive anchors to better handle multi-scale text using fewer anchors. The AutoFocus approach~\cite{najibi2018autofocus} predicts ``FocusPixels'', regions which are likely to contain small objects, and applies the multi-scaling process only on these regions, resulting in improvements in terms of runtime and memory efficiency. None of these approaches, however, propose an adaptive scaling scheme specifically suited for text instance segmentation.

\section{Proposed Approach}
The idea behind our approach is to first use a fast forward pass, over a downsampled image, to predict general text regions and their respective scales. These scales are then used to resize all the text regions into a uniform compact representation which is then forwarded through the same neural network to separate the words.

More specifically, our approach is composed of five steps as can be seen in Figure~\ref{fig:pipeline}. First, a single-scale detection network is applied to the given input image to detect general text regions and their scales. Secondly, regions are extracted using the segmentation mask. Thirdly, the information extracted from the first stage is used to create a compact knapsack containing only the regions of text, scaled to the same size as can be seen in Figure~\ref{fig:compacts}. Fourthly, the knapsack image is forwarded through the same single-scale detection network. Finally, a post-processing mechanism is used to extract the output image.

\begin{figure}
\centering

\includegraphics[width=0.11\textwidth]{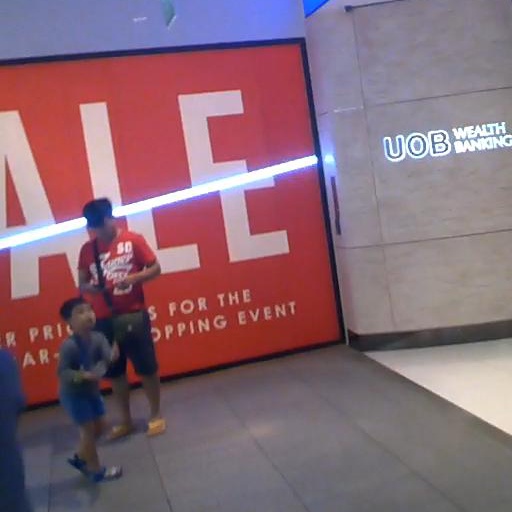}
\includegraphics[width=0.11\textwidth]{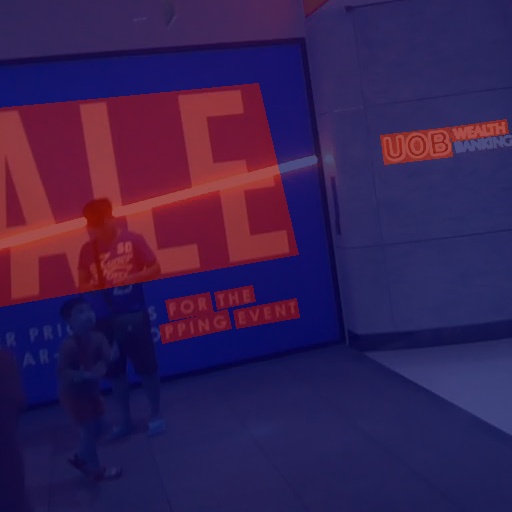}
\includegraphics[width=0.11\textwidth]{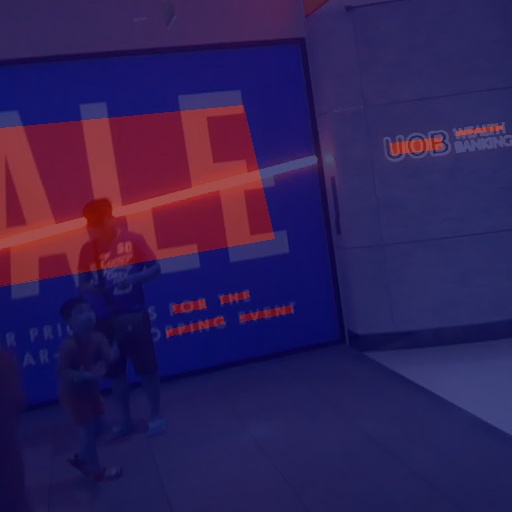}
\includegraphics[width=0.11\textwidth]{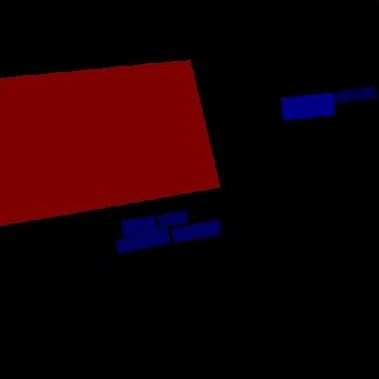}
\caption{Training data. From left to right, the input image, the segmentation map, the shrunk map, and the scale map. }
\label{fig:label}
\end{figure}

\subsection{The Single-Scale Method}
\begin{figure}[b]
    \centering
    \includegraphics[width=0.47\textwidth]{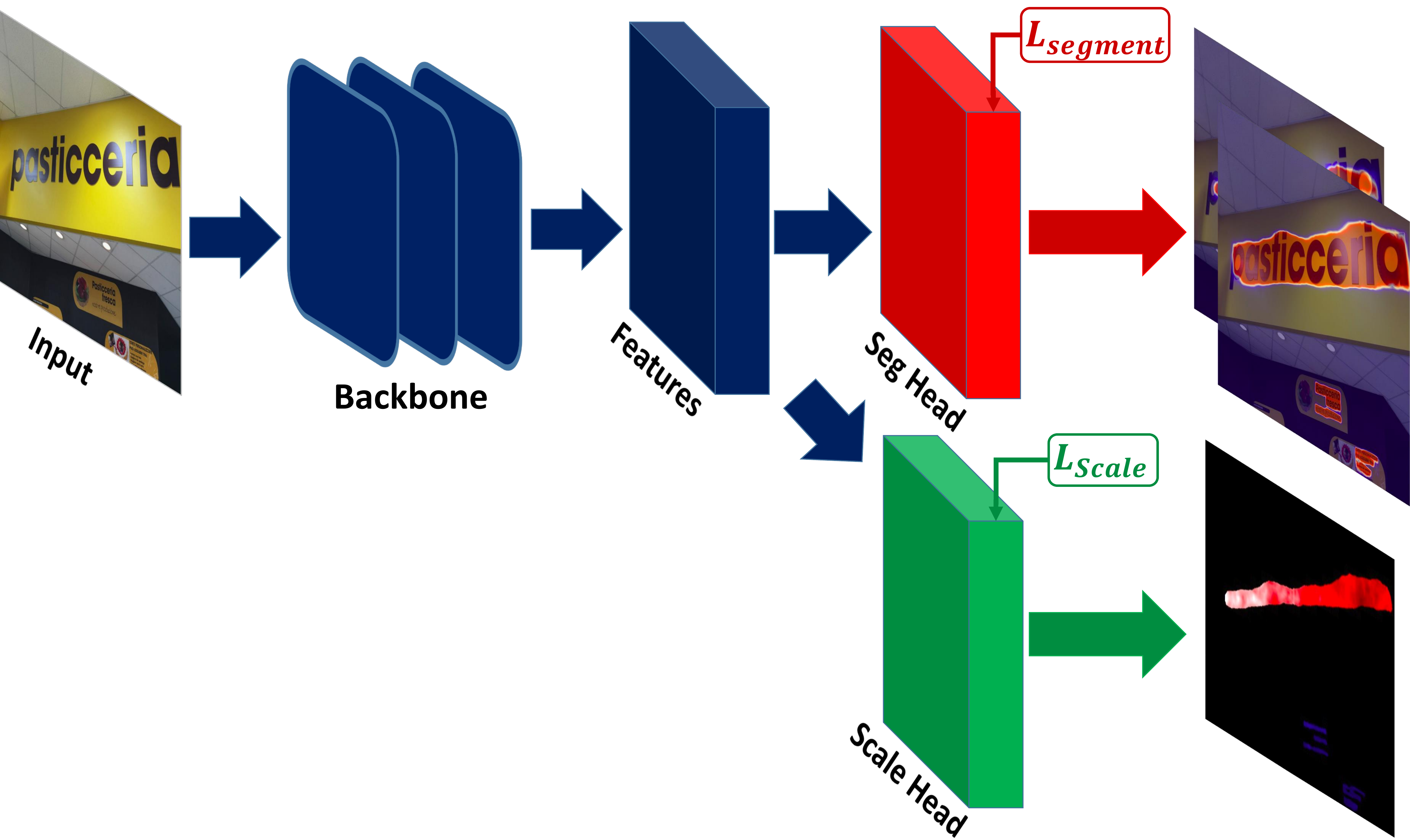}

    \caption{Proposed architecture. Image is fed through a convolutional backbone followed by segmentation and scale layers. }
\label{fig:proposed_archi}
\end{figure}

As shown in many recent works~\cite{deng2018pixellink,zhou2017east,wang2019shape}, using an architecture that predicts dense pixel-wise output maps grants us the flexibility to learn different forms of mappings, such as the Geometry Map of EAST~\cite{zhou2017east} and the Connectivity Map of PixelLink~\cite{deng2018pixellink}. Usually, the purpose behind these different representations is to allow effective extraction of well-separated bounding boxes from the output maps. While our adaptive scaling scheme can be used on top of many segmentation architectures, we chose to implement a new baseline as part of our approach. Figure~\ref{fig:proposed_archi} shows an overview of our single scale detection network. The backbone itself is discussed in Section~\ref{subsec:implement}.

Inspired by the recent work of PSENet~\cite{wang2019shape} and by the shrunk polygons used in the EAST method~\cite{zhou2017east} two output maps are learned. The first one is a simple text/no-text semantic map, while the second one is a shrunk map, where only the inner part of the polygon is classified as text, see Figure~\ref{fig:label}. The shrunk map is used for an improved distinction between close text instances.
As in~\cite{wang2019shape}, the shrunk map is created using the Vatti clipping algorithm~\cite{vati1992generic}, with the numbers of pixels to clip, $d$, defined as
\begin{equation}
    d=\frac{Area\left(P\right)\times\left(1-r^{2}\right)}{Perimeter\left(P\right)}.
\end{equation}
Here $P$ is the initial polygon and the scale ratio, $r$, is set to $0.4$.
Similar to~\cite{wang2019shape,he2019textcontournet}, the segmentation channels are trained using the dice-loss, which is defined as
\begin{equation}
  L(S,G)=1-\mbox{\ensuremath{\frac{2\sum_{x,y}\left(S_{x,y}\times G_{x,y}\right)}{\sum_{x,y}S_{x,y}^{2}+\sum_{x,y}G_{x,y}^{2}}}},
\end{equation}
where $S$ is the output segmentation map and $G$ is the ground-truth map. As the shrunk map is incorporated into full text map, the loss is applied only on the regions inside the text map. This has the effect of breaking down the learning process into two sub-problems, detecting text regions and localizing words inside each such region. The final loss is set as
\begin{equation}
   L_{segment}=0.5\cdot L_{c}+0.5\cdot L_{s},
\end{equation}
where $L_c$ is the dice-loss applied segmentation channel and $L_s$ is the dice-loss on the shrunk segmentation channel. Online Hard Negative Mining is applied on $L_c$ with a ratio of 3.
During post-processing, rotated rectangles are extracted directly from the shrunk map and expanded in accordance with the shrinking ratio. This results in a simple and efficient post-processing procedure. While the full segmentation map is not used to extract the text instances, it is helpful for better extraction of small text regions, as will be discussed below.
Note that unlike~\cite{wang2019shape} our method predicts only a single kernel map, thus avoiding the need for the expansion algorithm proposed there which is more helpful for irregular text shapes.

\subsection{Introducing The Scale Channel}

The proposed ``scale predictor'' is created by simply adding another output channel in the last convolutional layer, thus outputting a 3-channel image, containing both the segmentation masks and the scale.
The scale of each word is defined by finding the bounding rotated rectangle and taking its height, or smaller axis, as the scale of the entire word. The resulting label is shown in Figure~\ref{fig:label}. We found that height is a good choice for the scale value, as it is not affected by the number of characters and is closely related to the font size and the spacing between words. For inference we take the average scale inside each segment, weighted by the confidence of the segmentation map, as its scale.

The mathematical formulation of our scale prediction loss draws inspiration from anchor-based object detection methods, specifically~\cite{RCNN}. These methods perform bounding-box regression in order to transform default anchor boxes into tight object proposals, predicting 4 transformation parameters that are used for translation and scaling. The parameters are represented as log-space additive offsets which are equivalent to pixel-space multiplications. Having the scale represent multiplication suits our scenario very well, as it grants the same weight to different sized texts in the same image. That is, a 30-pixel high text that was predicted to be 15 pixels high would inflict the same loss as a 300 pixel high text that was predicted to be 150-pixels high. A simpler choice of directly predicting the text height in pixels would result in a drastic over-weighting of large texts, as their additive pixel difference would be much bigger than small texts. This results in the following formulation
 \begin{equation}
   \hat{s_{i,j}}=\log\left(\frac{s_{i,j}}{s_{ref}}\right),
 \end{equation}
where $s_{i,j}$ is the height of the word the pixel belongs to, and $s_{ref}$ is set to $25$.
Finally, a Smooth-L1 loss is applied over all regions labeled as text, which means that we do not need to define scale values for background regions. The $L_{scale}$ loss is defined as,

\begin{equation}
L_{scale}\left(\hat{s},\hat{s}_{gt}\right)=\begin{cases}
0.5\left(\hat{s}-\hat{s}_{gt}\right)^{2}, & \mbox{if }\left|\hat{s}-\hat{s}_{gt}\right|<1\\
\left|\hat{s}-\hat{s}_{gt}\right|-0.5 & \mbox{otherwise}
\end{cases},
\end{equation}
where $\hat{s}$ is the normalized scale, and $\hat{s}_{gt}$ is the ground truth normalized scale. $L_{scale}$ is added to $L_{segment}$ for joint training of segmentation and scale,
\begin{equation}
   L=L_{segment}+0.1\cdot L_{scale}.
\end{equation}
The values for $s_{ref}$ and the loss weights were set following empirical experiments.

\subsection{Refined Inference}

\begin{figure}[b]
  \vspace{-3mm}
  \centering
  \setlength\tabcolsep{0.1pt}
  \begin{tabular}{@{}ccccc@{}}
      \includegraphics[width=0.11\textwidth]{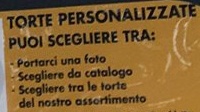}&&
      \includegraphics[width=0.11\textwidth]{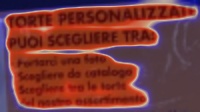}&
      \includegraphics[width=0.11\textwidth]{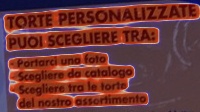}&
      \includegraphics[width=0.11\textwidth]{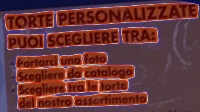}
      \tabularnewline
      \includegraphics[width=0.11\textwidth]{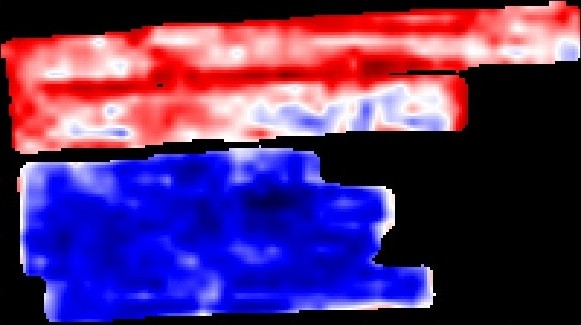}&
      \includegraphics[width=0.0075\textwidth]{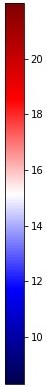}&
      \includegraphics[width=0.11\textwidth]{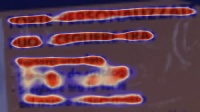}&
      \includegraphics[width=0.11\textwidth]{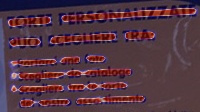}&
      \includegraphics[width=0.11\textwidth]{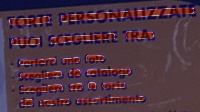}
      \tabularnewline
      &&
      \includegraphics[width=0.11\textwidth]{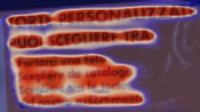}&
      \includegraphics[width=0.11\textwidth]{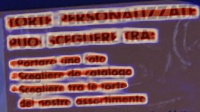}&
      \includegraphics[width=0.11\textwidth]{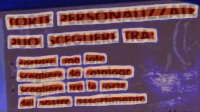}
      \tabularnewline
       & & $\times0.5$ scale & $\times1$ scale & $\times4$ scale \tabularnewline
      \end{tabular}
  \caption{Segmentation outputs under different scales. Each column presents the segmentation map followed by the shrunk segmentation map and the average map. Input image and scale map are shown in the first column.}
\label{fig:scaling}
\end{figure}

The first forward pass over the downsampled image is used to retrieve general text regions, without fine separation between small words, alongside their respective predicted scale.
Every text region, or ``blob'', is then extracted from the original image and resized to the desired scale, which is set as $1.5s_{ref}$.
To efficiently process the extracted blobs, all regions are packed together to create a compact knapsack representation. This is done using the Maximal Rectangles Best Short Side Fit algorithm~\cite{jylanki2010thousand}. The knapsacks are then passed through our network to create the refined segmentation result, from which the final rotated rectangles are extracted.

While text regions can be extracted directly from the shrunk segmentation map, using an averaged map of the two segmentation channels for the first stage results in better performance. This can be attributed to the fact that while the shrunk map gives better separation, it misses some of the smaller text regions, as shown in Figure~\ref{fig:scaling}.

\begin{figure}
\centering

\includegraphics[width=0.15\textwidth]{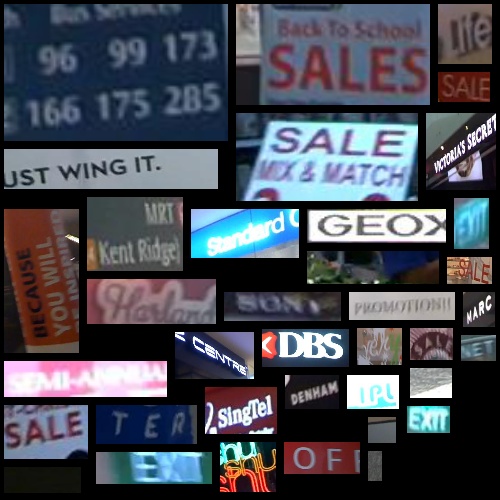}
\includegraphics[width=0.15\textwidth]{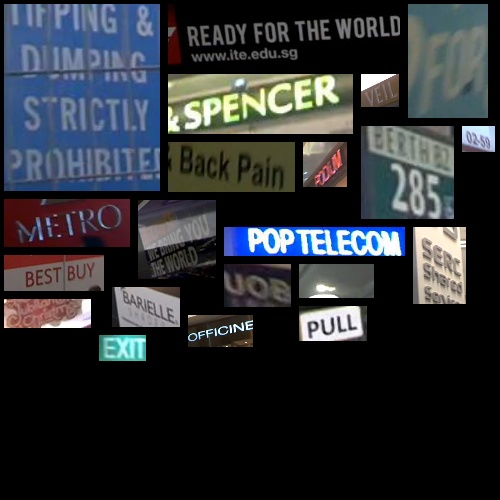}
\includegraphics[width=0.15\textwidth]{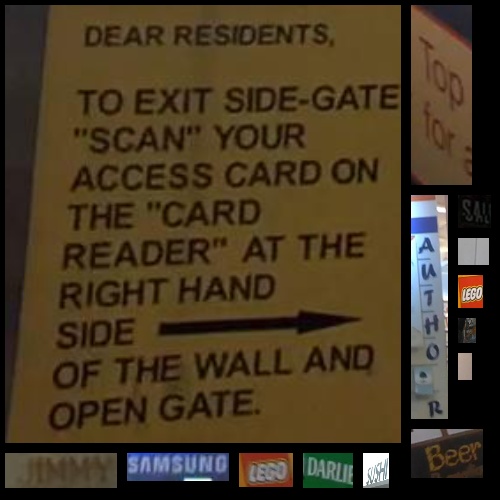}
\caption{Artificial knapsacks created for augmentation.}
\label{fig:knapsacks}
\vspace{-5mm}
\end{figure}

Note that while the knapsack images create a compact representation, they are inherently different from the natural images used for training, which can cause the network results to degrade. To mitigate that we simply propose to add a Knapsack Augmentation to the training process. This is done by randomly taking regions of text from different images, resizing them to our reference scale, and packing them using our packing scheme. Some generated knapsacks are shown in Figure~\ref{fig:knapsacks}.

\vspace{-2mm}
\section{Experiments}

Here, we evaluate the proposed baseline and adaptive scaling scheme and compare it to fixed scaling approaches. In all experiments, our single scale method is denoted as $\mbox{Ours}$, while our two-stage solution is denoted as $\mbox{Ours}^{+}$. Methods are evaluated on three different long side scales $720$, $1024$ and $1440$, which are denoted as $S$, $M$ and $L$ accordingly. Finally, $\mbox{Ours}_{L}$ denotes that a method was used with an input image resized to scale $L$.
 Experiments are conducted on the ICDAR15 and ICDAR17 benchmarks.
\subsection{Datasets}\label{subsec:data}

\paragraph{ICDAR15} The ICDAR15 Competition on Robust Reading~\cite{karatzas2015icdar} is a standard benchmark for detecting oriented text in-the-wild. The benchmark is composed of 1,500 images taken using a Google Glass sensor, where 1,000 images are used for training and 500 for testing. Text instances might be rotated and are tagged as quadrilaterals.

\paragraph{ICDAR17 MLT} The ICDAR17 Competition on Multi-Lingual Scene Text Detection~\cite{nayef2017icdar2017} is a large scale benchmark for text detection in multiple languages. The benchmark contains 7,200 training images and 9,000 test images taken from a diverse set of scenes. The benchmark is deemed challenging both due to the variability in text location and scale, and the need to recognize, and separate, words in different languages.

\subsection{Implementation Details}
\label{subsec:implement}
For our backbone, we investigate both an FPN module~\cite{lin2017feature}, which is a widely used architecture for semantic segmentation, and the ESPNet architecture from~\cite{mehta2018espnet}. For the FPN module, a pretrained ResNet-50~\cite{he2016deep} backbone is used, with an additional feature fusion mechanism, as in~\cite{wang2019shape}. Compared to the lightweight ESPNet, FPN is relatively heavy in terms of runtime and memory efficiency, a distinction that can help understand the effect of the underlying architecture on our method.

For training data, we use both the ICDAR15 and ICDAR17 MLT training images, where both datasets are balanced during training so that every batch is approximately evenly split. A standard augmentation pipeline is used during training,
 composed of the following steps (1) A photometric distortion process as in~\cite{liu2016ssd} (2) Aspect ratio distortion, where the height is scaled by a uniform random factor in the range of $[0.6,1.4]$  (3) Random scaling by a factor of $\left\{ 0.5,1,2,3\right\}$ (4) Rotation by an angle between $-10^{\circ}$ and $10^{\circ}$ (5) Random cropping of $640\times640$ pixels around a labeled text instance, and finally, (6) mirroring is randomly applied with a probability of $0.3$, where the cropped image padded as needed.
The FPN module is trained with a batch size of $12$ using stochastic gradient descent (SGD) with weight decay of $5\cdot 10^{-4}$ and Nesterov momentum of $0.99$. The network is trained for $180\cdot 10^3$ iterations where the initial learning rate is $1\cdot 10^{-3}$ and is decayed by a factor of $0.1$ every $60\cdot10^3$ iterations. ESPNet is trained with a batch size of $16$ using the ADAM~\cite{kingma2014adam} solver, where the initial learning rate is $1\cdot 10^{-3}$ and is decayed by a factor of $0.94$ every $10\cdot 10^3$ steps. Training was performed on 4 NVIDIA M60 GPUs, where a single one was used for evaluation.

\subsection{Benchmark Results}
\begin{table}
  \setlength{\tabcolsep}{1.5pt}
  \centering
  \begin{tabular}{@{}lllllll@{}}
  \toprule
  Method & Recall & Precision & F-Score & Forward & Process & FPS \\ \midrule
  $\mbox{PSENet}_{S}$   & 46.46\% & 72.01\% &  56.48\%    &  \multicolumn{1}{c}{$66ms$}        & \multicolumn{1}{c}{$35ms$}           &    \multicolumn{1}{c}{$9.9$}   \\
  $\mbox{PSENet}_{L}$   & 80.79\% &  83.65\%&    82.19\%   &   \multicolumn{1}{c}{$247ms$}      &           \multicolumn{1}{c}{$149ms$} &  \multicolumn{1}{c}{$2.5$}   \\
  $\mbox{CRAFT}_{S}$  & 60.13\% & 79.30\% &  68.40\%     &   \multicolumn{1}{c}{$77ms$}     & \multicolumn{1}{c}{$11ms$}      & \multicolumn{1}{c}{$11$}          \\
  $\mbox{CRAFT}_{L}$   & 80.50\% & 84.96\% &    82.67\%   &   \multicolumn{1}{c}{$301ms$}            &    \multicolumn{1}{c}{$45ms$}         &   \multicolumn{1}{c}{$2.9$}        \\ \midrule
  $\mbox{Ours}_{S}$   & $58.97\%$ & $82.16\%$  & $\color{blue}68.67\%$  &   \multicolumn{1}{c}{$66ms$}       &  \multicolumn{1}{c}{$10ms$}    & \multicolumn{1}{c}{$\color{blue}13$}       \\
    $\mbox{\textbf{Ours}}^{+}_{S}$       & $78.52\%$ & $83.60\%$  & $\color{blue}80.98\%$ & \multicolumn{1}{c}{$82ms$}        &    \multicolumn{1}{c}{$16ms$}        &  \multicolumn{1}{c}{$\color{blue}10$}   \\
  $\mbox{Ours}_{L}$    & $80.60\%$ & $84.72\%$  & $\color{green}82.61\%$      &  \multicolumn{1}{c}{$244ms$}       &   \multicolumn{1}{c}{$17ms$}         & \multicolumn{1}{c}{$\color{green}3.8$}    \\
  $\mbox{\textbf{Ours}}^{+}_{L}$       & $83.05\%$ & $85.35\%$  & $\color{green}84.19\%$ & \multicolumn{1}{c}{$262ms$}        &    \multicolumn{1}{c}{$27ms$}        &  \multicolumn{1}{c}{$\color{green}3.5$}   \\
   \bottomrule
  \end{tabular}

  \caption{Results on the ICDAR15 Benchmark. ``Forward'' is the GPU runtime, while ``Process'' includes all the processing done on CPU.}
  \label{tb:icdar15}
\end{table}

\begin{figure*}
  \centering
  \setlength\tabcolsep{1.5pt}
  \begin{tabular}{@{}ccc@{}}
    & x$6$ pixel reduction \vspace{-4.5mm}&  \tabularnewline
        \includegraphics[width=0.33\textwidth]{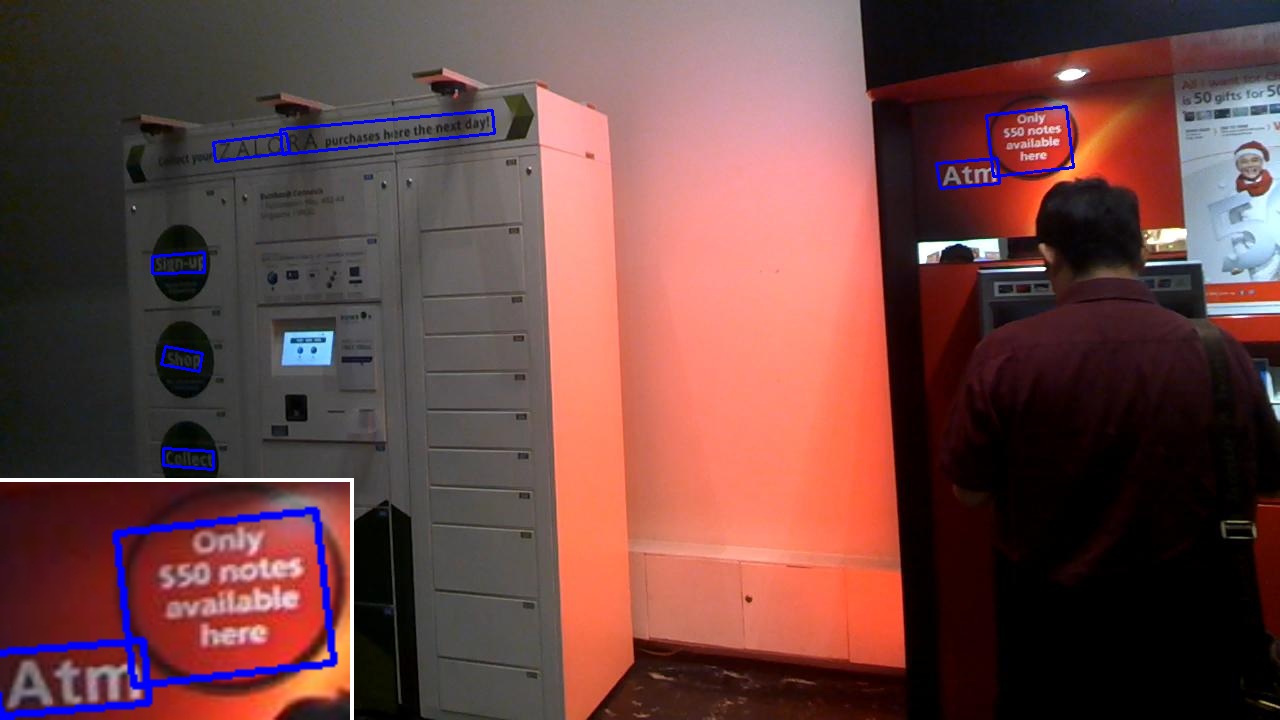}&
      \includegraphics[width=0.24\textwidth]{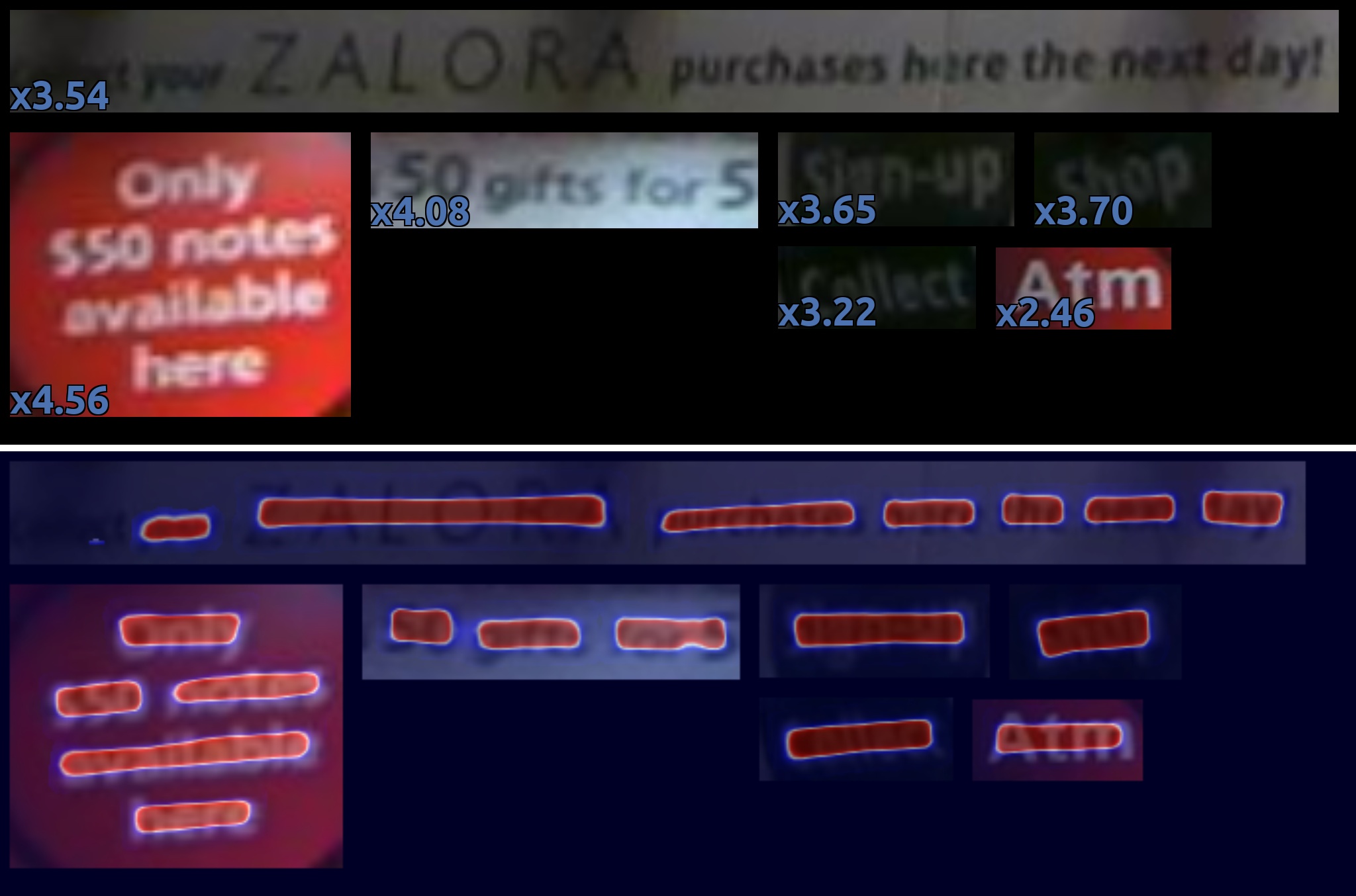}&
      \includegraphics[width=0.33\textwidth]{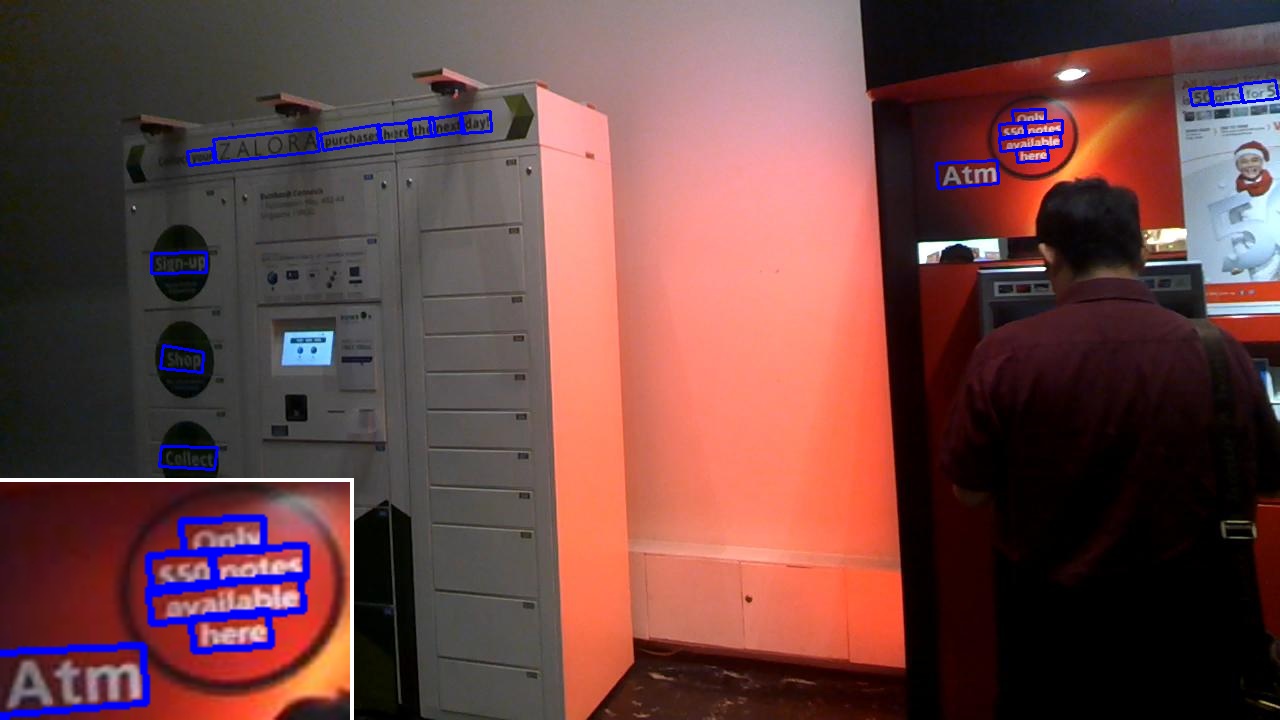}\vspace{3mm}\tabularnewline
      & x$6$ pixel reduction \vspace{-6.5mm}&  \tabularnewline
      \includegraphics[width=0.33\textwidth]{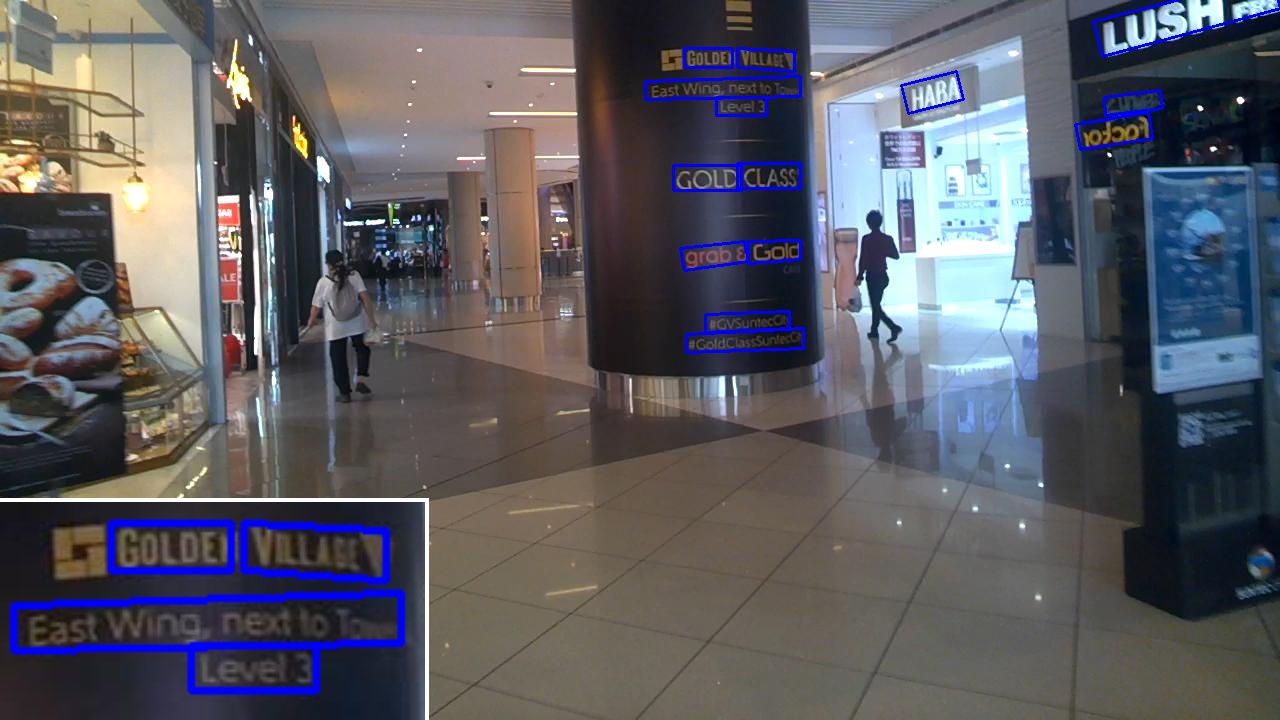}&
      \includegraphics[width=0.23\textwidth]{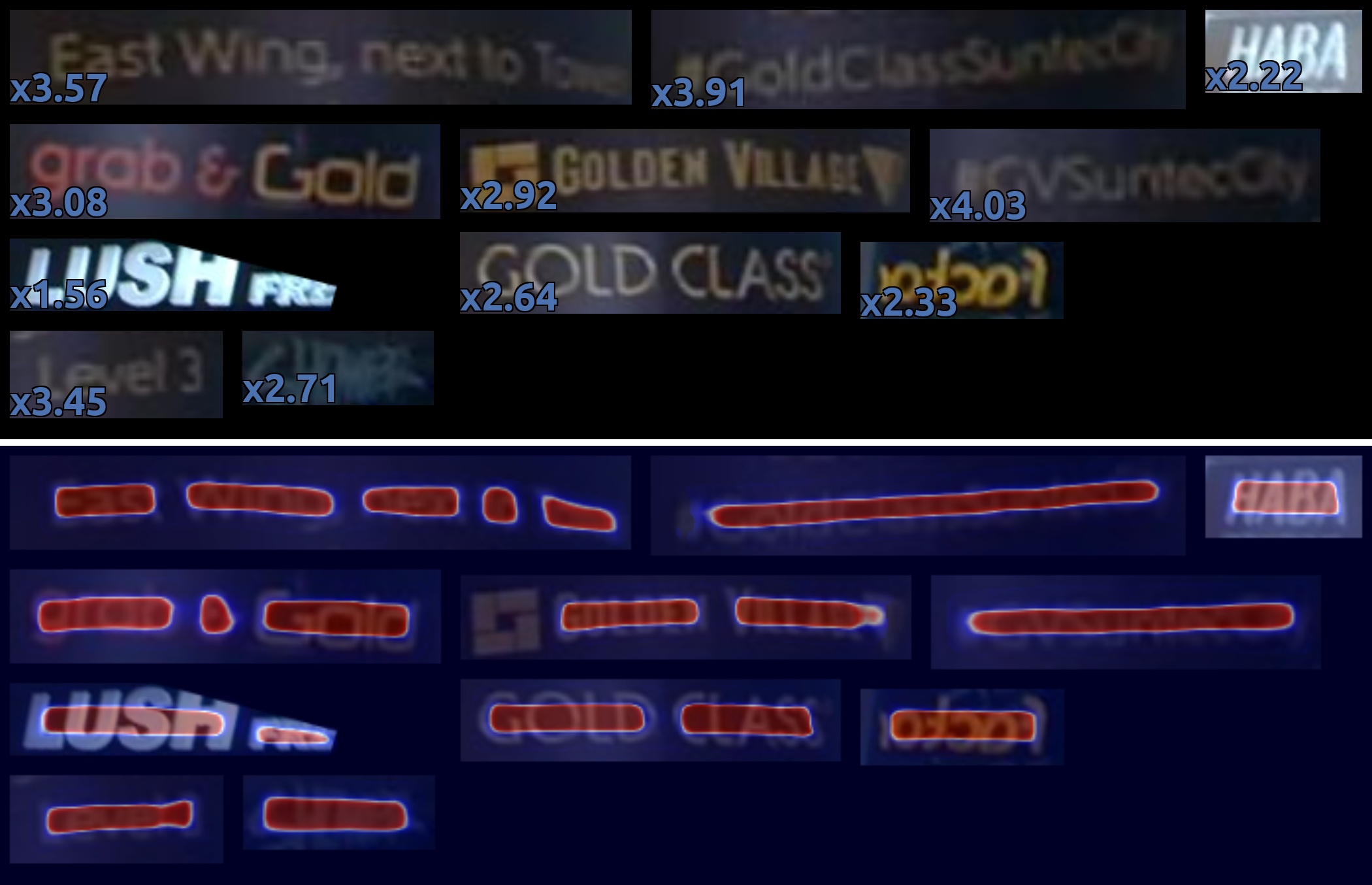}
      &
      \includegraphics[width=0.33\textwidth]{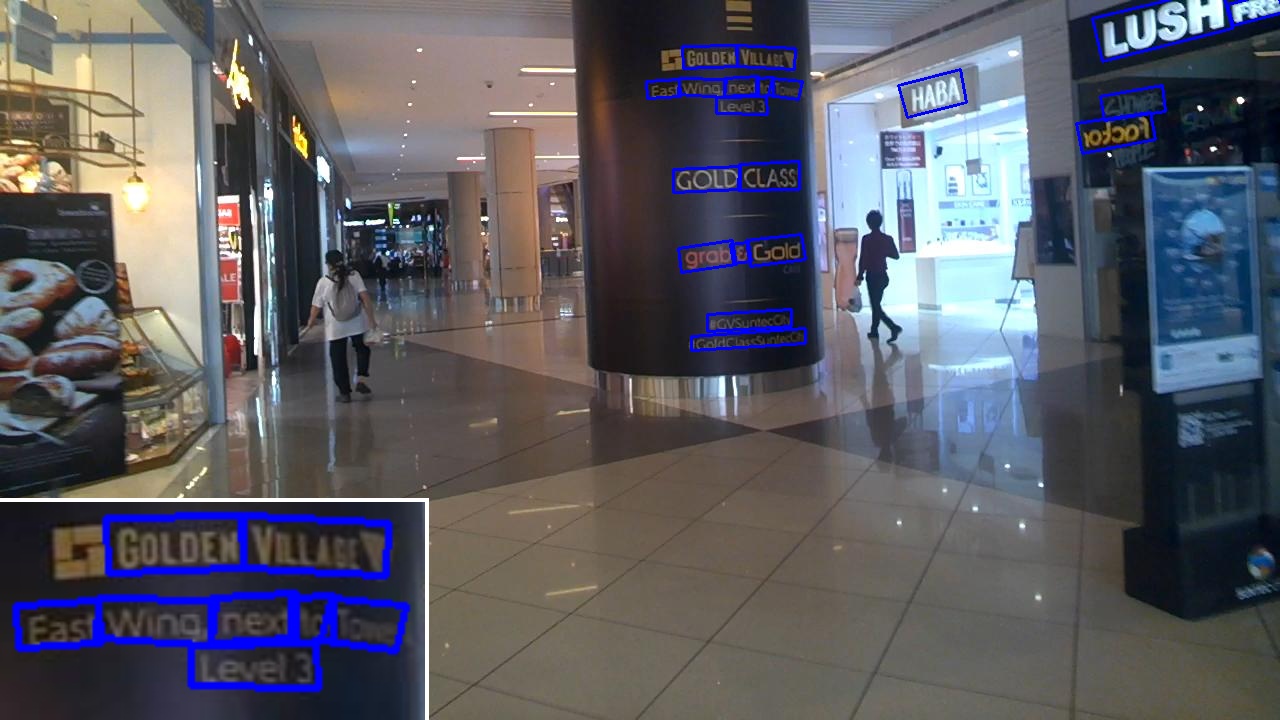}\tabularnewline
      $\mbox{\textbf{Ours}}_{S}$ & \textbf{Generated knapsack} & $\mbox{\textbf{Ours}}^{+}_{S}$ \tabularnewline
      \end{tabular}
  \caption{Results on the ICDAR15 benchmark. Scaling factor on every region is shown with respect to the input size for $\mbox{\textbf{Ours}}_{S}$. The pixel reduction is calculated as the ratio between the original image size and the created knapsack. }
\label{fig:icdar15}
\end{figure*}

The proposed approach is evaluated on the ICDAR15 and 17 benchmarks alongside PSENet~\cite{wang2019shape} and CRAFT~\cite{baek2019character}. Results on PSENet and CRAFT were produced using the official implementations, but with the same input scale as our method, where their hyperparameters were reselected for these scales. This allows us to accurately compare the performance of the method itself decoupled from its input size. Note that all experiments were run on the same NVIDIA M60 machine using Pytorch~\cite{paszke2017automatic}.

\paragraph{ICDAR15 Results} Table~\ref{tb:icdar15} presents the results on the ICDAR15 benchmark.
 One can see that, when using the same fixed scales, our single scale method achieves comparable results to state-of-the-art segmentation methods. It is also clear that the $L$ scale is consistently more accurate but is also significantly less efficient. Next, looking at the proposed adaptive scaling scheme, $\mbox{Ours}^{+}_{S}$, one can see that indeed our method is able to stay close to $\mbox{Ours}_{S}$ in terms of runtime while producing results of much larger scales. This shows that \textbf{our method is able to somewhat mitigate the inherent trade-off between runtime and accuracy.} Note that Figure~\ref{fig:results_graph} shows these results as a ``runtime vs F-Score'' graph for better visualization. Qualitative results are presented in Figure~\ref{fig:icdar15}, showing both the output of  $\mbox{Ours}_{S}$ and  $\mbox{Ours}^{+}_{S}$ alongside the generated knapsacks. This also shows the efficiency of using a compact representation compared to the original image.

   \begin{table}[b]
     \setlength{\tabcolsep}{1.5pt}
     \centering
     \begin{tabular}{@{}lllllll@{}}
     \toprule
     Method & Recall & Precision & F-Score & Forward & Process & FPS \\ \midrule
     $\mbox{CRAFT}_{S}$  & 42.56\% & 74.46\% &  54.16\%     &   \multicolumn{1}{c}{$91ms$}            &    \multicolumn{1}{c}{$16ms$}         &   \multicolumn{1}{c}{$9.3$}          \\
     $\mbox{CRAFT}_{L}$   & 61.68\% & 78.80\% &    69.20\%   &   \multicolumn{1}{c}{$351ms$}            &    \multicolumn{1}{c}{$65ms$}         &   \multicolumn{1}{c}{$2.4$}        \\ \midrule
     $\mbox{Ours}_{S}$   & $39.36\%$ & $73.10\%$  & $\color{blue}51.17\%$  &   \multicolumn{1}{c}{$83ms$}       &  \multicolumn{1}{c}{$10ms$}    & \multicolumn{1}{c}{$\color{blue}10.9$}       \\
       $\mbox{\textbf{Ours}}^{+}_{S}$       & $51.93\%$ & $76.27\%$  & $\color{blue}61.79\%$  & \multicolumn{1}{c}{$106ms$}        &    \multicolumn{1}{c}{$27ms$}        &  \multicolumn{1}{c}{$\color{blue}7.48$}   \\
     $\mbox{Ours}_{L}$    & $56.77\%$ & $74.21\%$  & $\color{green}64.33\%$      &  \multicolumn{1}{c}{$311ms$}       &   \multicolumn{1}{c}{$35ms$}         & \multicolumn{1}{c}{$\color{green}2.88$}    \\
     $\mbox{\textbf{Ours}}^{+}_{L}$       & $61.82\%$ & $72.94\%$  & $\color{green}66.92\%$ & \multicolumn{1}{c}{$329ms$}        &    \multicolumn{1}{c}{$64ms$}        &  \multicolumn{1}{c}{$\color{green}2.54$}   \\
      \bottomrule
     \end{tabular}
     \caption{Results on the ICDAR17 benchmark.}
 \label{tb:icdar17}
   \end{table}

\paragraph{ICDAR17 Results}
 Table~\ref{tb:icdar17} shows that the results on the challenging ICDAR17 benchmark. Our method uses the same settings, but with a knapsack scale of $1.8s_{ref}$. One can see that CRAFT, which was pretrained on the SynthText dataset~\cite{gupta2016synthetic}, is stronger than our baseline in the multi-lingual scenario. Still, our adaptive scheme proves successful. That is $\mbox{Ours}^{+}_{S}$ is significantly more accurate than $\mbox{Ours}_{S}$, getting close to the performance of $\mbox{Ours}_{L}$ while staying twice as fast. Note that both methods do not reach state-of-the-art results in these configurations, which usually require extremely large scales. PSENet originally rescales each image by a factor of two, while the CRAFT method uses a long side of $2560$ pixels, resulting in significantly higher processing time.
 Quantitative results on ICDAR17 are shown in Figure~\ref{fig:icdar17}.

\begin{table}
\centering
  \setlength{\tabcolsep}{2pt}
    \begin{tabular}{@{}llllll@{}}
    \toprule
      $\mbox{Backbone}_{1}$ &   $\mbox{Backbone}_{2}$ & Recall & Precision & F-Score & \multicolumn{1}{c}{FPS}  \\ \midrule
       \multicolumn{1}{c}{FPN}  &\multicolumn{1}{c}{---} & 58.98\% & 82.16\% & 68.66\% & 13.13 \\
       \multicolumn{1}{c}{ESPNet}  &\multicolumn{1}{c}{---} & 46.17\% & 74.17\% & 56.91\% & 43.68 \\
       \multicolumn{1}{c}{FPN}  &\multicolumn{1}{c}{FPN} & 78.52\% & 83.60\% & 80.98\% & 10.05 \\
       \multicolumn{1}{c}{ESPNet}  &\multicolumn{1}{c}{ESPNet} & 66.49\% & 78.64\% & 72.06\% & 28.09 \\
       \multicolumn{1}{c}{FPN}  &\multicolumn{1}{c}{ESPNet} & 72.17\% & 78.93\% & 75.40\% & 11.01 \\
       \multicolumn{1}{c}{\textbf{ESPNet}}  &\multicolumn{1}{c}{\textbf{FPN}} & \textbf{70.72\%} & \textbf{84.57\%} & \textbf{77.03\%} & \textbf{23.62} \\
    \bottomrule
    \end{tabular}

\caption{ Backbone hybrids. Different combinations of the FPN and ESPNet backbones on the ICDAR15 benchmark are evaluated. $\mbox{Backbone}_{1}$ is the backbone used for the initial segmentation and $\mbox{Backbone}_{2}$ is the backbone used over the knapsacks.}
\vspace{-5mm}
\label{tb:archi}
\end{table}

\begin{figure}
    \centering
    \includegraphics[width=0.47\textwidth]{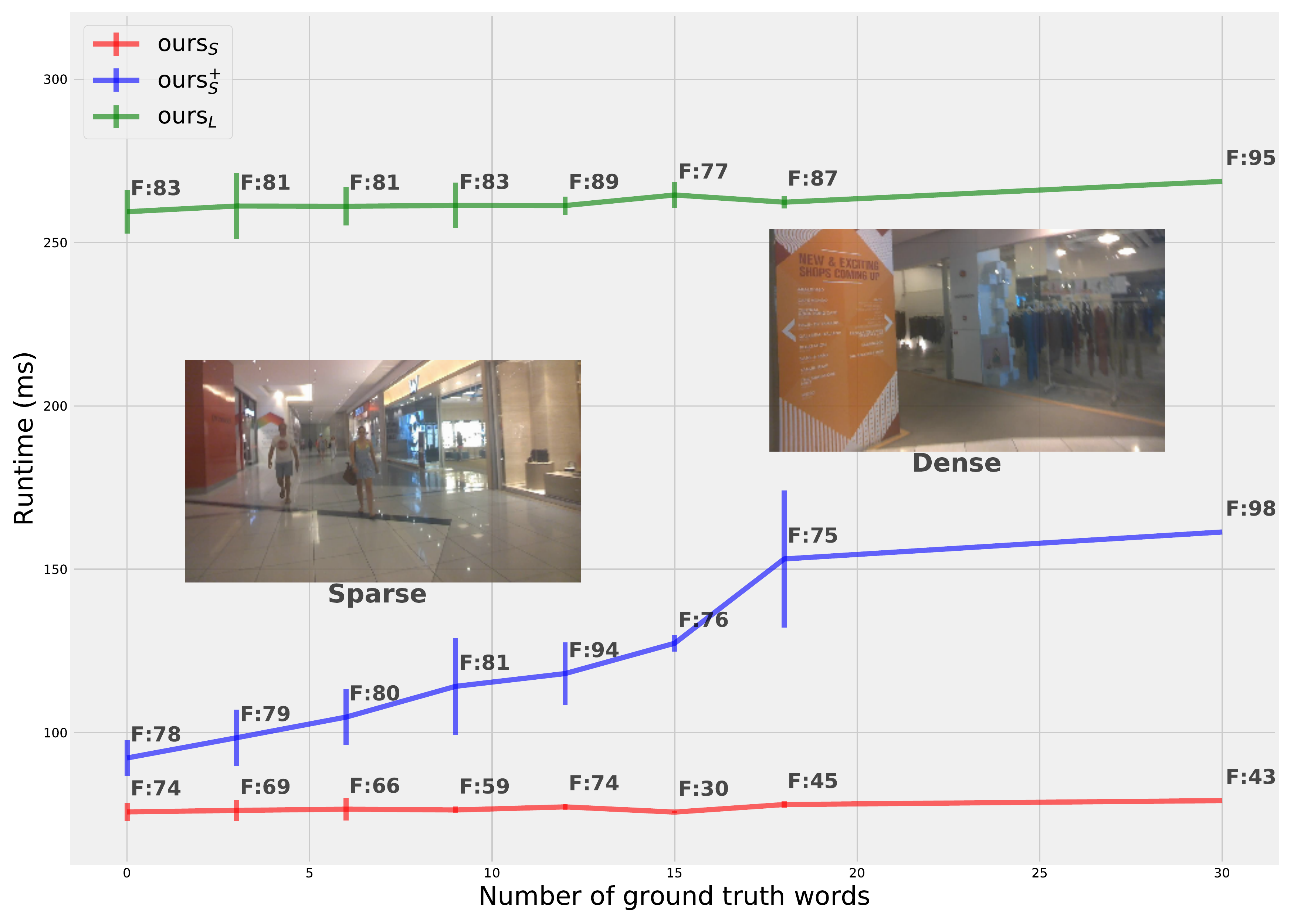}

    \caption{Runtime by number of words in ICDAR15. Average F-Score is shown for each point. Sample images for each region are also shown. }
\label{fig:runtime_words}
\end{figure}

\subsection{Additional Analysis}
\paragraph{On choosing the backbone} We now turn to investigate how using different backbone architectures affects our results by comparing the behavior of ESPNet and FPN when used in the first or second segmentation stages. Table~\ref{tb:archi} shows the results of our method in the $S$ scale over the ICDAR15 benchmark where different backbones were used. One can see that while FPN is indeed more accurate, it is significantly slower than the ESPNet architecture and that both methods benefit from our adaptive solution. In the rest of our experiments we chose to use FPN as our baseline due to its increased accuracy. An interesting configuration is the usage of ESPNet for the first stage, which requires only coarse segmentation, and FPN in the second stage for the refined localization. This composition is almost twice as fast than the single-stage FPN while also increasing the F-Score by $8\%$.

\begin{figure*}
  \centering
  \setlength\tabcolsep{1.5pt}
       \vspace{11mm}
  \begin{tabular}{@{}ccc@{}}
    & x$12$ pixel reduction \vspace{-19mm}&  \tabularnewline
      \includegraphics[width=0.33\textwidth]{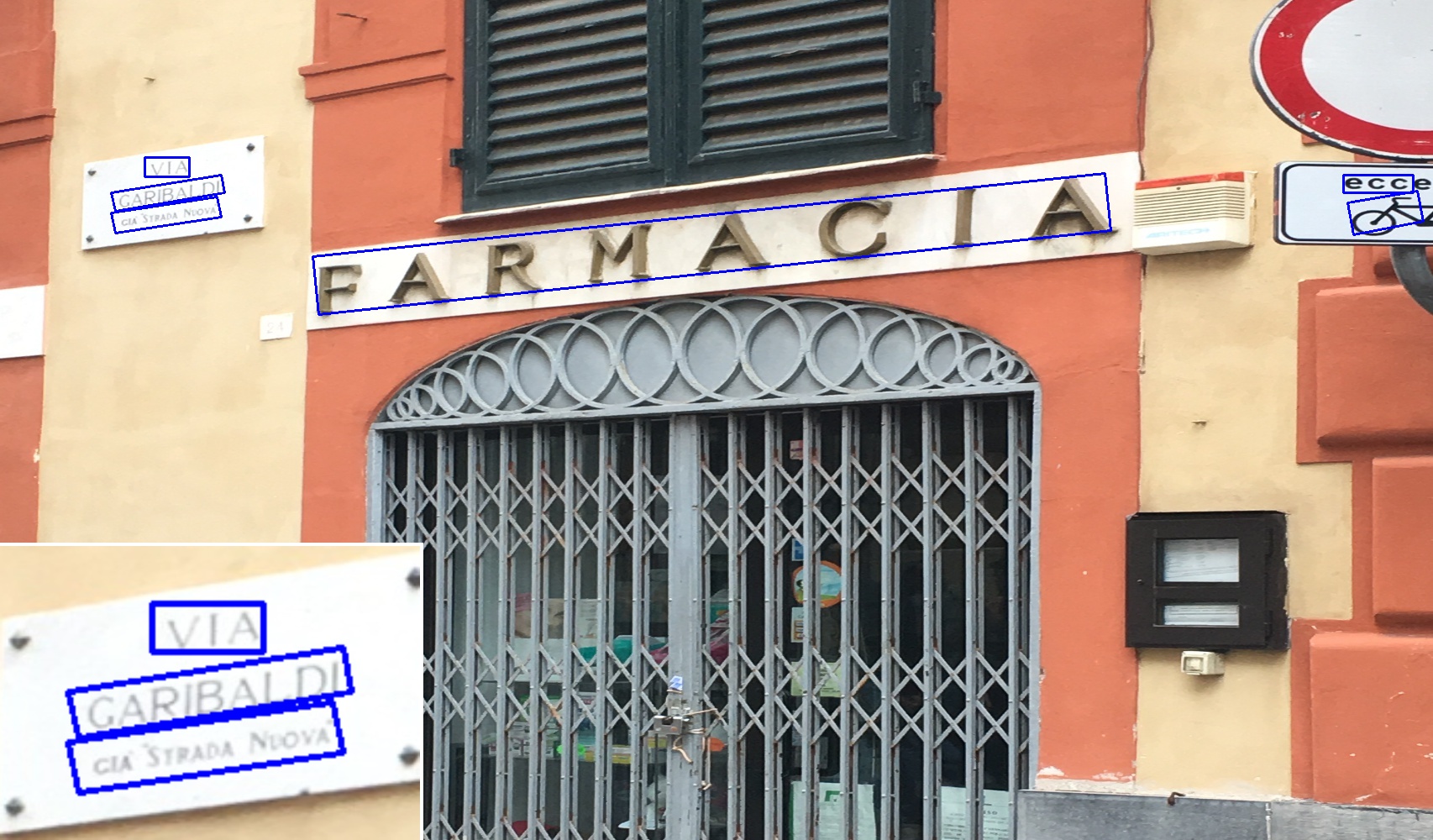}&
      \includegraphics[width=0.215\textwidth]{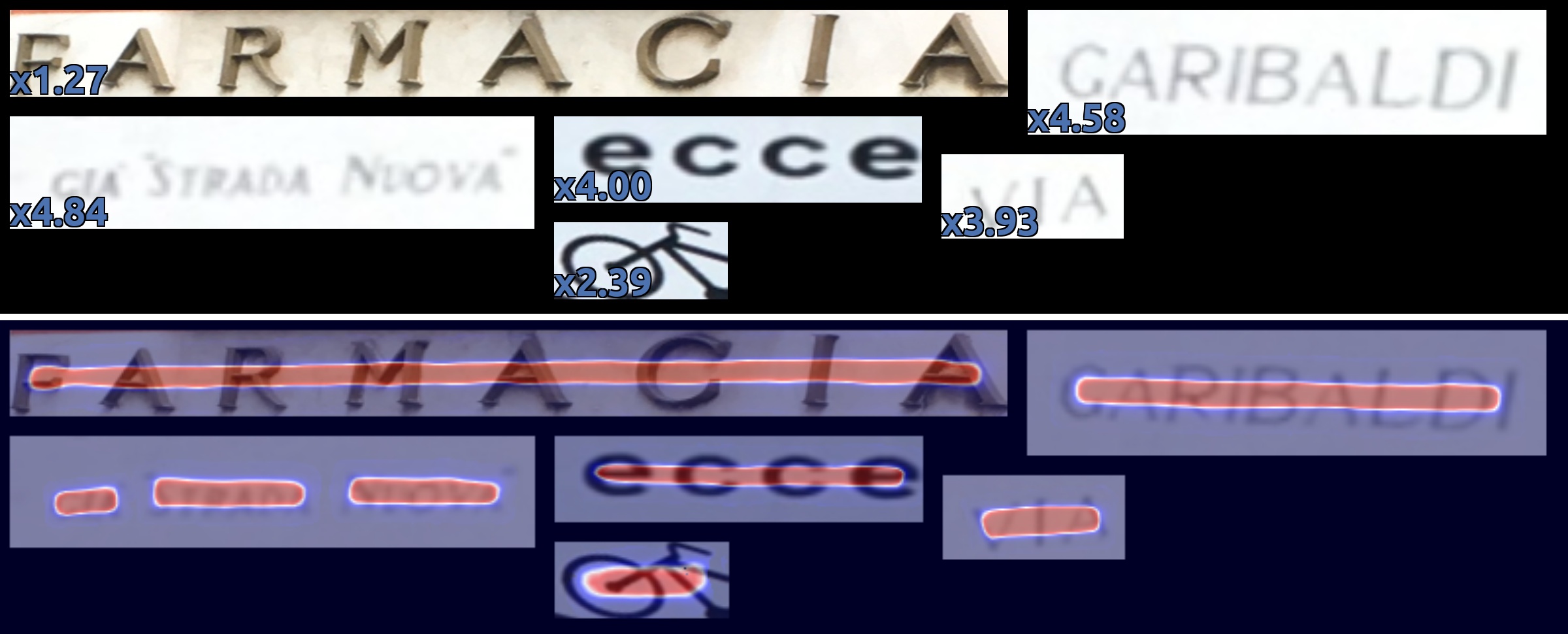}&
      \includegraphics[width=0.33\textwidth]{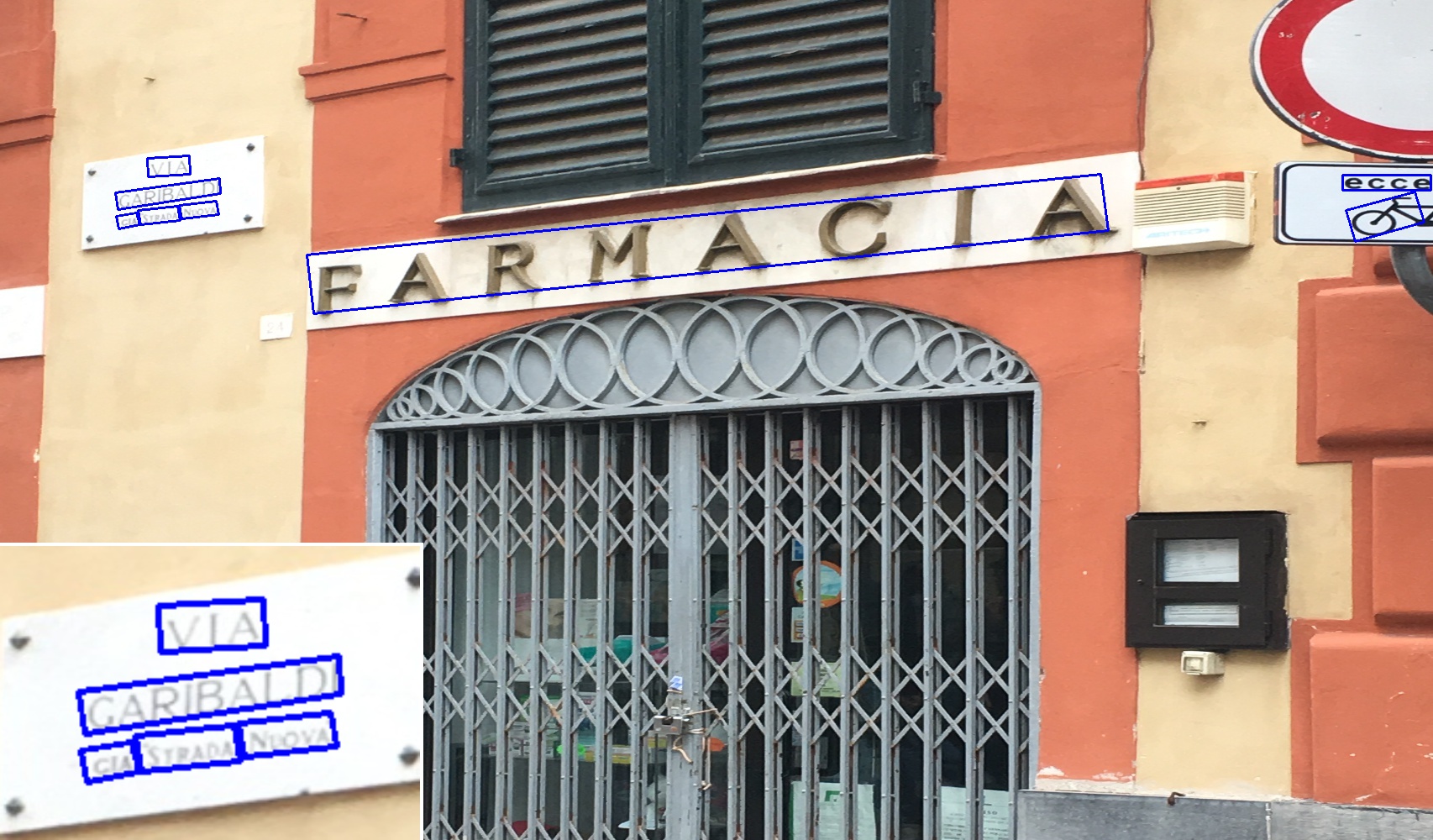}\vspace{2mm}\tabularnewline
      & x$7$ pixel reduction  \vspace{-3.7mm}&  \tabularnewline
      \includegraphics[width=0.33\textwidth]{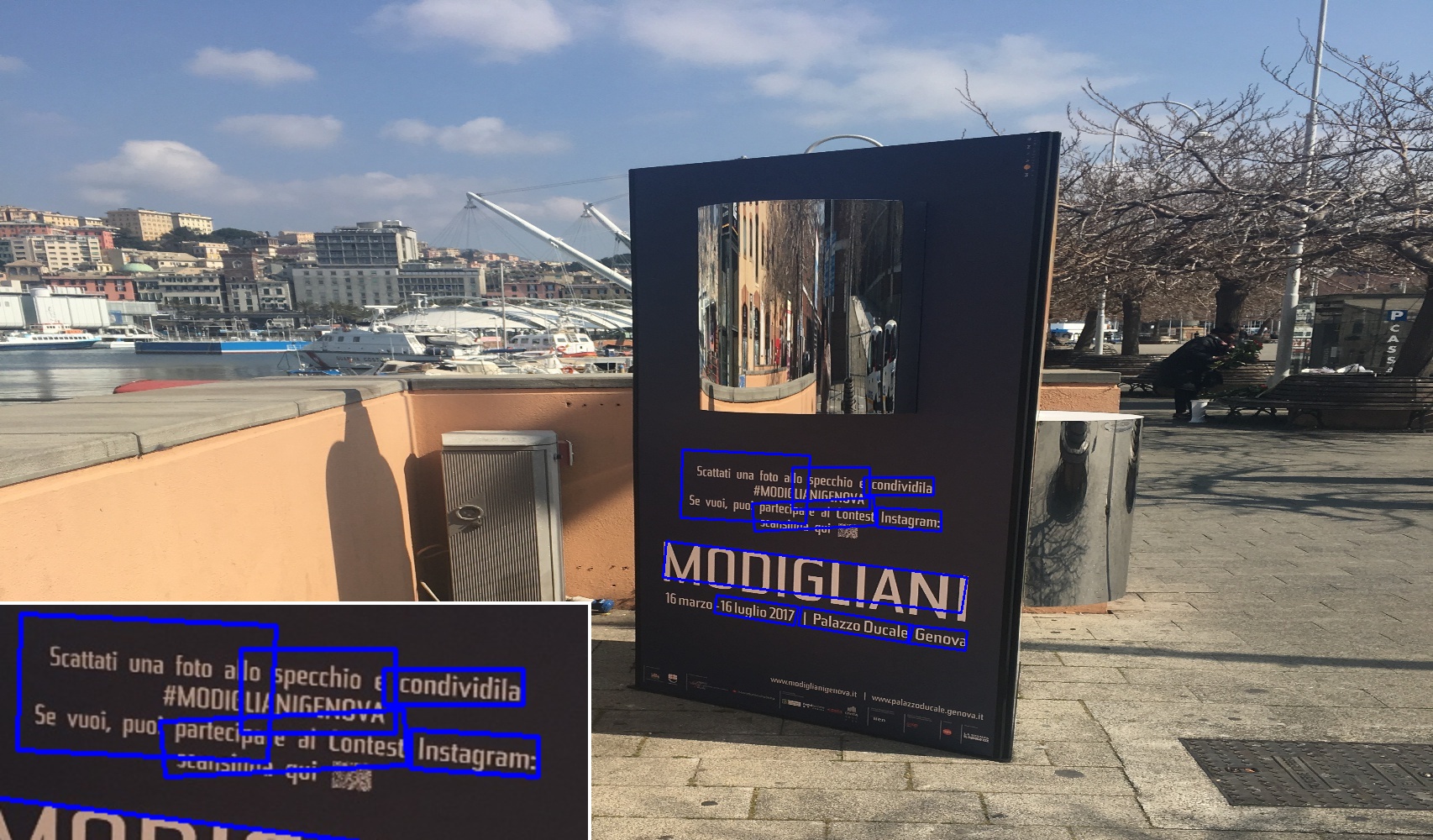}&
      \includegraphics[width=0.19\textwidth]{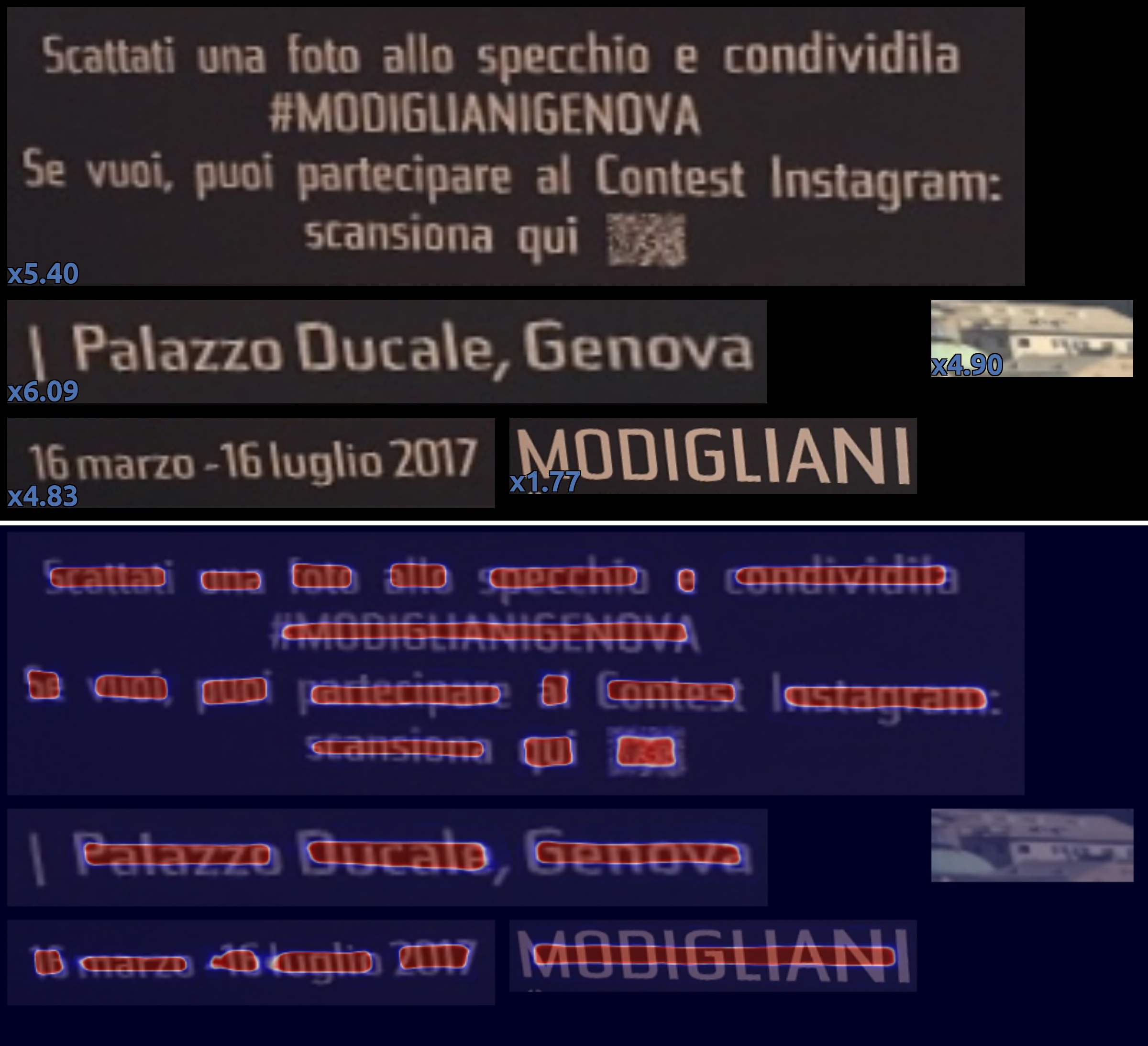}&
      \includegraphics[width=0.33\textwidth]{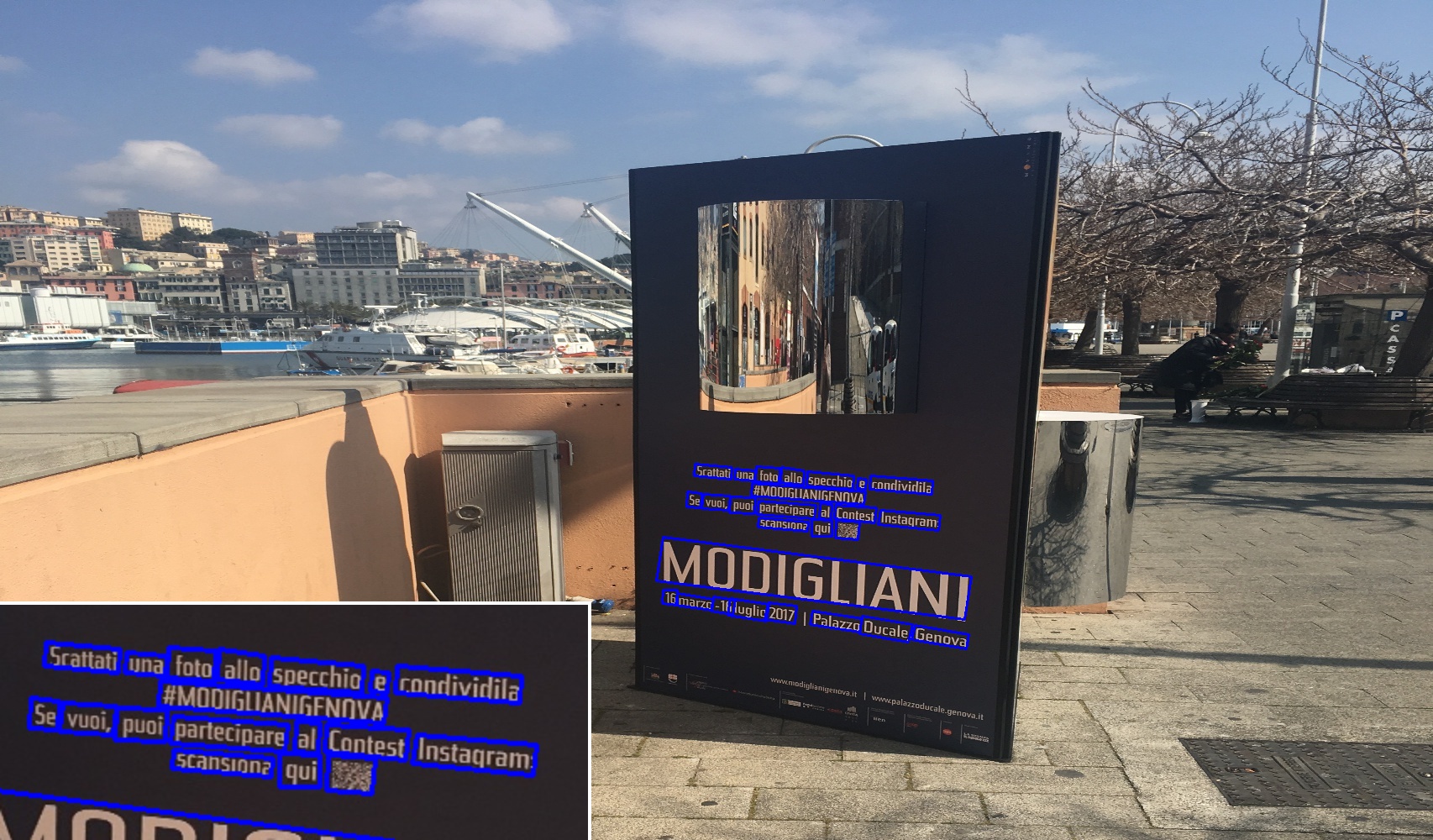}\tabularnewline
      $\mbox{\textbf{Ours}}_{S}$ & \textbf{Generated knapsack} & $\mbox{\textbf{Ours}}^{+}_{S}$ \tabularnewline
      \end{tabular}
  \caption{Results on the ICDAR17 MLT benchmark. Some images were cropped for better visualization.}
\label{fig:icdar17}
\vspace{-2mm}
\end{figure*}

\paragraph{On text sparsity and runtime} One of the interesting properties of the proposed method is its adaptive runtime. Where fixed scaling schemes would use approximately the same runtime for different images, our method adaptively changes the processing time according to the amount of text in the image. Figure~\ref{fig:runtime_words} shows the average processing time as a function of the number of words in the image, on the ICDAR15 benchmark. As expected one can see that while  $\mbox{Ours}_{S}$ is constantly fast and $\mbox{Ours}_{L}$ is constantly slower, $\mbox{Ours}^{+}_{S}$ processes sparse images with almost no overhead while staying faster than $\mbox{Ours}_{L}$ even on the more dense images.

\begin{figure}
    \centering
    \includegraphics[width=0.47\textwidth]{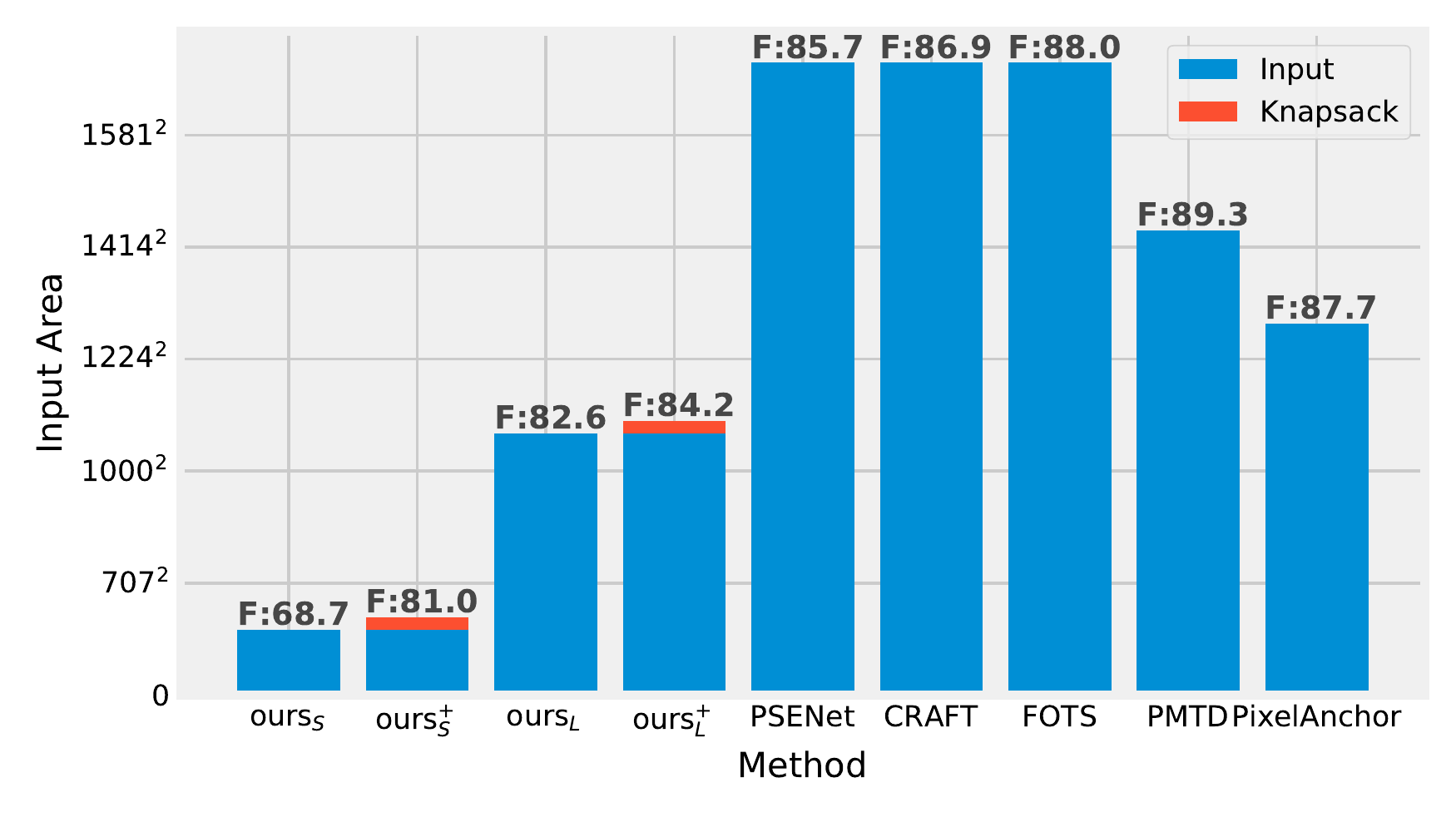}
\vspace{-4mm}
    \caption{Input pixels for ICDAR15. For each method, the amount of pixels fed into the network is shown, for our method we include the area of compact representation as well. F-scores are shown on top of each bar according to the reported results of PSENet~\cite{wang2019shape}, CRAFT~\cite{baek2019character}, FOTS~\cite{liu2018fots}, PMTD~\cite{liu2019pyramid} and Pixel-Anchor~\cite{li2018pixel}.}

\label{fig:results_bar}
 \vspace{-4mm}
\end{figure}

\paragraph{On processed pixels} While previous figures analyze our results in terms of runtime, another interesting evaluation is the number of input pixels fed to the network. While correlated with runtime, this analysis brings us another perspective as it is independent on the backbone and hardware used by the method.
In Figure~\ref{fig:results_bar} we show the overall input area fed into the network of our different configurations alongside state-of-the-art methods, where the best reported single scale configuration is shown. Note how our adaptive method can significantly increase the F-score with only a small amount of extra area from the second pass. This can be attributed to the fact that indeed the knapsacks are significantly smaller than the original input image. One can see that our $\mbox{Ours}^{+}_{L}$ configuration is close to those reported by PSENet and CRAFT on $1260\times2240$ input images while processing fewer pixels. Interestingly, recent methods such as the Pyramid Mask Text Detector~\cite{liu2019pyramid} can get impressive results while processing $1080\times1920$ images. We believe that our adaptive scaling could be a powerful extension for these methods as well.

\section{Limitations}
As shown in our experiments, our approach achieves results that are on-par with larger scales while reducing the runtime. Still, we note that there are some limitations to the proposed approach. Mainly, when choosing an extremely small scale for the first stage, the baseline might fail to segment some of the smaller text regions, resulting in them not appearing in the final result. Furthermore our method builds upon the fact that text is usually sparse, for dense images such as documents or newspapers our method might not be as effective.

\section{Conclusion}
\vspace{-2mm}
We presented a novel adaptive scaling scheme for efficient text detection. Our approach uses a semantic segmentation network to detect coarse text regions while simultaneously predicting their scale. This information is then used to create a compact representation containing only the scaled text regions, from which refined word instances are extracted using an additional segmentation stage.
Our approach is shown to be a powerful alternative to fixed scaling schemes, achieving the accuracy of larger scales in a more efficient manner.

\vspace{-2mm}
\subsubsection*{Acknowledgments}
\vspace{-1mm}

The authors would like to thank Mr. Alon Palombo for his support and insights during the early stages of this research.

{\small
\bibliographystyle{ieee}
\bibliography{egbib}
}

\end{document}